# Playing optical tweezers with deep reinforcement learning: in virtual, physical and augmented environments

MATTHEW PRAEGER,[1,*,#] YUNHUI XIE,[1,#] JAMES A. GRANT-JACOB,[1,] ROBERT W. EASON[1,] AND BEN MILLS[1]

[1] *Optoelectronics Research Centre, University of Southampton, SO17 1BJ, UK.*
[#] *These authors contributed equally*
[*]*mattp@soton.ac.uk*

**Keywords:** optical tweezers, laser trapping, machine learning, reinforcement learning

**Abstract:** Reinforcement learning was carried out in a simulated environment to learn continuous velocity control over multiple motor axes. This was then applied to a real-world optical tweezers experiment with the objective of moving a laser-trapped microsphere to a target location whilst avoiding collisions with other free-moving microspheres. The concept of training a neural network in a virtual environment has significant potential in the application of machine learning for experimental optimization and control, as the neural network can discover optimal methods for problem solving without the risk of damage to equipment, and at a speed not limited by movement in the physical environment. As the neural network treats both virtual and physical environments equivalently, we show that the network can also be applied to an augmented environment, where a virtual environment is combined with the physical environment. This technique may have the potential to unlock capabilities associated with mixed and augmented reality, such as enforcing safety limits for machine motion or as a method of inputting observations from additional sensors.

## 1. Introduction

Neural networks have enabled a significant wave of innovation within the field of photonics, including developments in microscopy [1-3], optical fibers [4, 5], silicon photonics [6, 7], sensing [8-10] and modelling light-matter interaction [11-13]. Predominantly, these results have been based on the machine learning technique known as supervised learning, where a neural network is trained to predict an output for a given input, such as transforming an optical microscopy image (the input) into one with a particular type of staining or fluorescence (the output). Whilst supervised learning has proved to be extremely effective when the data is formed of input-output pairs, a different approach is needed for experimental control and optimization. Reinforcement learning [14-16] is a machine learning technique where the neural network learns by making a sequence of interactions with its environment, whilst being provided with a reward signal based on its success in achieving favorable outcomes. In other words, the neural network is free to explore the environment, via trial and error, to discover an optimal process for solving a specified problem.

A critical technique in reinforcement learning is the use of a virtual training environments (also known as a gyms [17]) to enable accelerated learning, where the neural network gains experience in the virtual training environment before it is applied to the physical environment (i.e. the real world). The key components of the gym concept, definition of the problem, the creation and subsequent training inside the virtual environment, and the application to the physical environment are shown in Fig. 1 a). The application of a gym means that the neural network training speed is dependent on computing power rather than the movement speed of physical equipment, whilst also providing other advantages such as protecting against equipment damage and alleviating material consumption. The challenge is therefore in creating a virtual environment that approximates the response of the physical experiment whilst providing a reward signal that accurately represents the consequences of equivalent actions made by the neural network in the physical environment.

Reinforcement learning in a virtual environment has recently become an important technique for robotic sample manipulation [18-20] and for collision avoidance applications such as virtual training of drones [21, 22] and autonomous driving [23]. Reinforcement learning has also recently produced human-beating performance at Atari [24], first person shooter [25, 26], strategy [27, 28], 3D multiplayer [29] and text-based [30-32] computer games, along with board games including chess, shogi, and Go [33-35]. The application of reinforcement learning to the field of photonics has recently intensified with example demonstrations including optimized laser welding [36], the control of mode-locking in an oscillator [37], the use of robotics to align an interferometer [38], perform wavefront correction [39], routing in optical transport networks [40] and coherent beam combination [41].

In this work we apply reinforcement learning at the interface of robotics and photonics to demonstrate non-contact microscale sample manipulation. The task for the neural network, as shown in Fig. 1 b), was to use the optical tweezers effect [42] to transport a microsphere from its initial location to a pre-defined target position without collision with other microspheres, and whilst having a limited field of view (equivalent to that observed in the physical setup). A neural network, which was trained in a virtual environment, was able to learn route finding and collision avoidance and apply this correctly when applied to the real-world physical environment. In addition, we demonstrate that it is possible to overlay additional virtual layers upon observations of the physical environment. This technique has the potential to unlock capabilities associated with mixed and augmented reality, such as adding safety limits for machine motion or for incorporating additional sensor inputs. This demonstration of reinforcement learning for control of light-matter interactions shows the potential for automation and optimization across a range of nano and microscale photonics, engineering, and bio-chemical tasks. In this manuscript, the physical environment and virtual environments are explained in sections 2 and 3, the virtual environment training process and demonstration is shown in sections 4 and 5, and the experimental results associated with the physical and then augmented environments are discussed in sections 6 and 7.

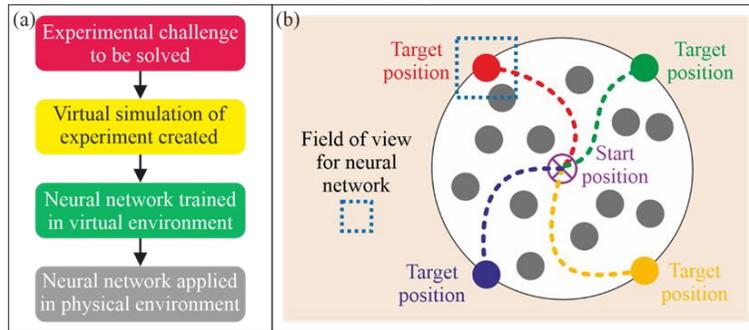

Fig. 1. a) Strategy for reinforcement learning through use of a virtual environment and b) concept for the task described here (showing trajectories to four different target positions).

## 2. The physical environment

The physical environment was based on an optical tweezers education kit (Thorlabs EDU-OT2/M). The beam from a laser diode operating at 658 nm with an output power of 40 mW (Thorlabs L658P040) was expanded and focused through a Zeiss, 63x, 0.8 N.A. microscope objective lens to generate the microsphere trapping force. The X and Y motorized stages each had a maximum travel of 12 mm, a maximum velocity of 2.6 mm/s, and a minimum step size of 0.05 µm but with backlash of up to 8 µm (which was partially mitigated in the control software). The motors were driven using Thorlabs KDC101 motion controllers commanded via custom python code that interfaced directly with the Thorlabs Kinesis drivers. The sample consisted of 5 µm diameter silicon dioxide microspheres in aqueous suspension (Sigma Aldrich 44054-5ML-F) which were diluted in distilled water to achieve the desired concentration (approximately 0.125% solids by mass). The liquid sample was sealed between a glass microscope slide and a cover slip using an adhesive gasket. The stages moved the sample relative to the laser trap and microscope objective lens, hence, producing a relative movement between the trapped microsphere and all other microspheres.

As shown in Fig. 2, in the physical environment the neural network observed the processed microscope image of the sample and was provided with a vector to the target location, and then responded by updating the stage velocities accordingly. Specifically, at each timestep, the camera image was processed to produce a binary image that was equivalent to that created in the virtual environment. Image segmentation was achieved using a sequence of conventional image processing methods including Gaussian blur, image standardization, adaptive thresholding, erosion and dilation, where both upper and lower thresholds were applied for tolerance against small variations in the microscope focus that could cause the microsphere centers to appear either darker or lighter than the mid-tone background. The center positions of particles in the segmented image were determined using a contour-based method [43]; this information was used to render a gym-equivalent image for the neural network. At each timestep the current stage positions were used to calculate a vector in the direction of the target (this was scaled relative to the initial displacement so that it also encoded the current distance from the target). The magnitude of this vector was clipped to avoid values which were not encountered during training, where the maximum displacement was 1000 pixels. The resulting vector was provided to the neural network as a numerical observation (note that the neural networks is never directly told the current stage positions or the target position). With the present hardware, neural network actions were operated approximately four times per second, limited by the communication speed of the motion controllers.

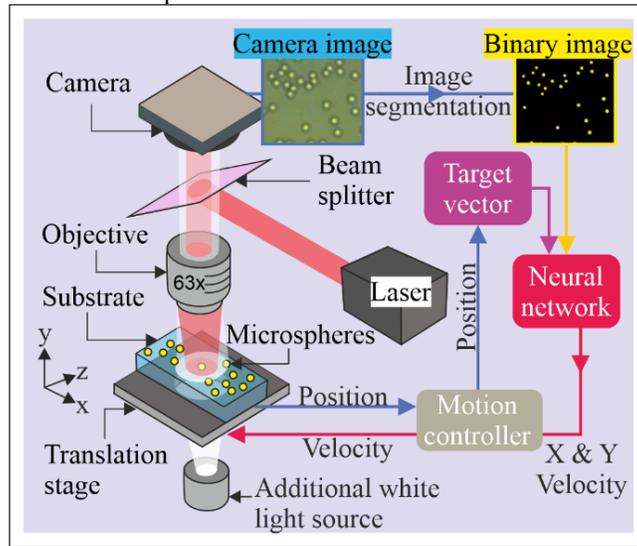

Fig. 2. Schematic of physical environment.

### 3. The virtual environment

As shown in Fig. 3, the virtual gym environment was programmatically designed to simulate the response of the real-world laser tweezers experiment. When the gym environment was initialized at the start of each episode, the target location was set, a single microsphere was initialized within the laser tweezers trap and a random number of free microspheres (between 500 and 1000) were randomly positioned within the virtual environment. At each timestep of the gym environment, each free microsphere was given a random acceleration, subject to a maximum allowed velocity, in order to simulate Brownian motion. Importantly, whilst the numerical values for microsphere positions and velocities were contained within the gym environment (as part of the world state), these were not made available to the neural network, and instead the microsphere coordinates were used to render a binary image that simulated the field of view of the optical tweezers microscope. To provide a match to the optical tweezers apparatus in the physical environment, an image resolution of 1280 x 1024 pixels was used and microsphere diameters were set at 54 pixels (corresponding to the 5 µm diameter size of the silicon dioxide microsphere used). To reduce computational requirements, the virtual microscope images were rendered at a lower resolution via a shrink factor of 4x resulting in an image with 320 x 256 pixels. This reduced-resolution, binary image was sent to the neural network at each timestep (we term this the image observation).

The gym environment also provided the neural network with a vector that denoted the displacement of the target location relative to the current stage position (we term this the numerical observation). At each timestep, the neural network was given the image and numerical observations and it used this information to select an action (velocity values for the X and Y stages). The world state was then calculated for the start of the next timestep, where the laser focus was assumed to remain central in the rendered camera image, with the stage motions simulated by translating all microspheres except for the one designated as trapped. The episode was immediately ended if multiple microspheres entered the laser trap region (i.e. a collision occurred) or if the laser trap was successfully maneuvered to the target location.

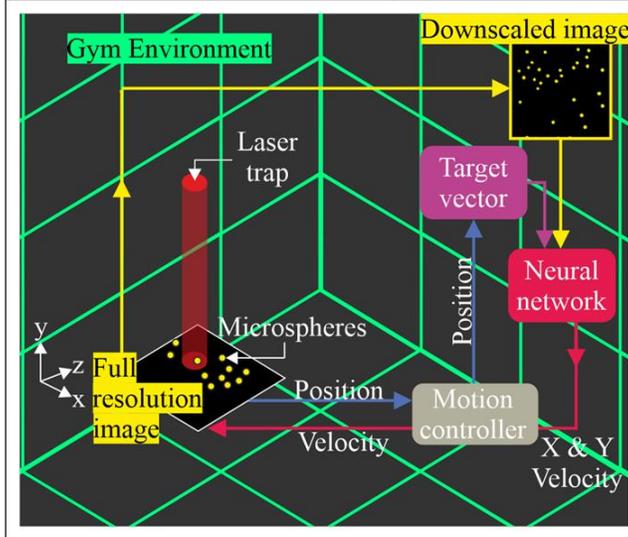

Fig. 3. Schematic of the virtual gym used for testing the trained neural network.

## 4. Training in the virtual environment

The reinforcement learning was carried out using the Twin Delayed Deep Deterministic Policy Gradient (TD3) algorithm [44]. As shown in Fig. 3, the neural network received a binary image (corresponding to the sample) and a two-component vector that pointed in the direction of the target relative to the current X and Y stage positions. The network design follows that used by [45], where the number of filter layers is doubled each time the feature map size is halved [46, 47]. The 256 x 320 binary image passed through three convolutional layers which, respectively, produced layer outputs with dimensions 64 x 80 x 32, 32 x 40 x 64 and 32 x 40 x 64, each with batch normalization and RELU activations. This was followed by three fully connected layers of size 512, 64 (+ numerical inputs), and 64, before producing an output vector of size 2 corresponding to the intended velocities for the X and Y axes. In the case of the actor the vector describing the direction towards the target was appended at the second fully connected layer, in the case of the critic both the target vector and the action were appended. Training was conducted in an episodic manner with a maximum episode length of 1000 time steps. At the start of each episode, the target was positioned at a randomly chosen angle, 1000 pixels away from the trap. During the rollout of an episode, image and numerical states were stored in a first-in, first-out memory buffer with up to 1050000 quadruples [48]. Each time an episode reached a terminal state (collision, success, or time limit) the update procedure commenced, where a batch of 32 state transitions was sampled randomly (with a uniform distribution) from the memory buffer and network parameters were updated accordingly. Analysis of the neural network training is shown in Fig. 4, where the effects of image resolution, reward shape and discount factor are explored.

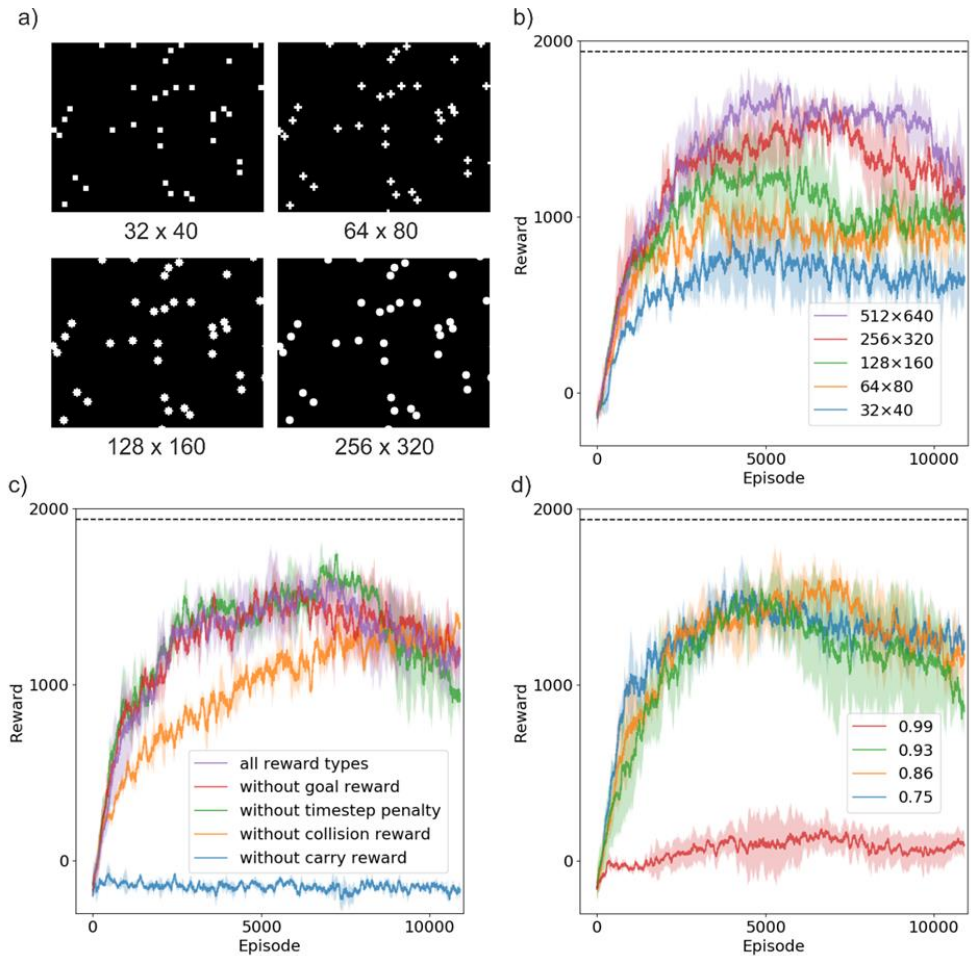

Fig. 4. Analysis of training in the virtual environment. a) Shows examples of binary images rendered at different resolutions. b-d) Show the effect on reward of different: b) image resolutions, c) reward combinations, d) discount factors. In b-d the horizontal dashed line shows the theoretical maximum reward, solid curves and shaded areas show the mean and standard deviation of a 100 episode, rolling-average of reward (evaluated over three training runs).

Fig. 4 a) shows the effect of resolution on the appearance of microspheres when rendered as a binary image, where increasing the resolution reduces the pixelization of the microspheres. Fig. 4 b) shows the 100 episode, rolling average of reward as a function of episode number, for binary image inputs with different resolutions. (Three training trials were performed for each case and so the solid curves in Fig. 4 b) and d) represent the mean reward whilst the shaded areas show the standard deviation of reward). Whilst higher resolution produced a higher reward, computational limitations resulted in a maximum practical resolution of 256 x 320, which was the resolution used throughout this work. Such resolution is comparable to other demonstrations of state observations in the literature, such as 84 x 84 x 4 [24], 64 x 64 x 9 [49], and 60 x 108 x 3 [26], where the third dimension arises from the use of separate RGB channels and from stacking of images from multiple timesteps.

The set of rewards that are provided to the neural network during training is critical in encouraging the preferred behavior, as a poor choice of reward structure can leave loopholes that can be exploited, leading to unexpected and undesirable behavior patterns that nonetheless yield high rewards. In this work, the rewards were i) a universal negative reward of -1 for each timestep, ii) a carry reward equal to the number of pixels moved towards the target (positive) or away from it (negative), iii) a collision reward of -100 for hitting another microsphere, and iv) a goal reward of +1000 for reaching the target location. (The values of these rewards were selected based on intuition and have not been optimized – systematic adjustment of these as hyperparameters could undoubtedly improve performance but would exceed our current computation capacity). The universal negative reward was intended to exert time pressure on the neural network. The carry reward was intended to encourage movement towards the target location. The collision reward punished the neural network for collisions with the randomly moving microspheres, which also terminated the episode. The goal reward was intended to reinforce the behavior of carrying a microsphere in the laser tweezers trap to the target location. To quantify the effectiveness of each of the reward components, training runs were carried out in which each of the components was eliminated one at a time, as shown in Fig. 4 c). When the carry reward was eliminated the network failed to learn the task. This is attributed to the fact that the carry reward provides feedback to the network at every time step to encourage movement towards the target. The second largest effect was observed when the collision reward was eliminated, which had the effect of reducing the learning rate. Whilst removal of the universal timestep negative reward and the goal reward appear to have only a minimal effect on performance, the highest performance was shown to occur when all four reward types were active.

The discount factor and discounted rewards [16] allow large single rewards to be distributed over a sequence of actions leading up to that reward, rather than reinforcing only the final action. The effectiveness of the terminal goal and collision rewards used here relies heavily upon the use of discounted rewards. For example, the direction of the last step that takes the microsphere into the target zone may be relatively unimportant, however, the continuous progress towards the target, over the previous hundred or so steps, is critical. The numerical value of the discount factor effectively determines the number of steps into the future that the network attempts to predict future rewards. Fig. 4 d) shows a comparison of the training performance for different discount factors, in the case where all reward components are included. The often-used value of 0.99 [44] was found to be inappropriate for this task and achieved very poor performance; this is attributed to the fact that higher discount factors require prediction of rewards over a longer horizon, where a large fraction of that trajectory may be outside of the limited field of view of the microscope. Consequently, the variance of these predictions is greater, which slows the rate of convergence and in severe cases causes the agent to completely fail to learn the task. As shown in Fig. 4 d) as the value of discount factor was reduced, the rate of convergence increased, with the highest average reward being achieved with a discount factor of 0.86. In all cases, the reward reaches a peak at around 6000 epochs, after which the performance declines. The precise cause of this effect is difficult to determine, however, it may represent an instance of catastrophic forgetting [50] for which one contributing factor can be the finite buffer size. However, as the models were saved periodically during training, the model that yielded the highest reward was applied to the physical environment.

## 5. Demonstration in the virtual environment

Once trained, the neural network was evaluated in the gym environment for a range of starting conditions. An example of a single trajectory is shown in Fig. 5, where the microsphere of interest and its associated trajectory is red, other microspheres are green circles, and the target position is a blue ring. Fig. 5 a) shows the trajectory of the trapped microsphere, along with the movement of the other microspheres. Fig. 5 b) shows the extent of the field of view (which does not include the entire path to the target position) and the starting positions of the free microspheres within that field of view. Fig. 5 c) shows the environment state at the end of the episode. As observed, the neural network is clearly capable of avoiding collisions whilst continuing to progress towards the target position, as the field of view is updated.

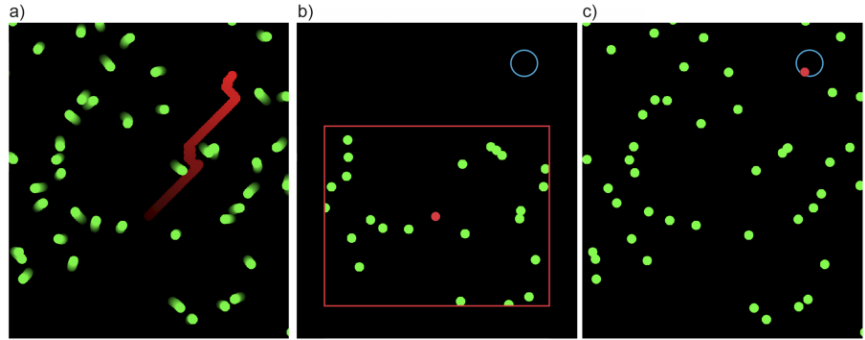

Fig. 5. Demonstration of the virtual gym environment for dynamic microspheres, showing a) the chosen trajectory and movement of microspheres during the episode b) the field of view and target position at the start of the episode and c) the environment state at the end of the episode.

To provide a more general demonstration of the capability of the neural network inside the gym, Fig. 6 shows twelve trajectories, each under identical starting conditions but with different target positions. Here, the simulated Brownian motion of the free microspheres was kept the same for each of the twelve trials, and therefore all trajectories can be overlaid on a single figure. Where the red and green paths appear to cross in the figure, this corresponds to a different time step position rather than a collision. Fig. 6 a) shows the trajectory paths for all microspheres, Fig. 6 b) shows the initial environment state, with associated field of view and target positions, and Fig. 6 c) presents the environment state after completion of all trajectories. Although the action space of the environment was continuous, the neural network was observed to frequently output actions corresponding to the combination of maximum positive or negative stage velocities for both axes. This is attributed to the combination of the time based reward (which encourages maximum velocity in many cases) and the use of the tanh hyperbolic function as the network activation function (which can result in saturation of actions towards their upper and lower bounds).

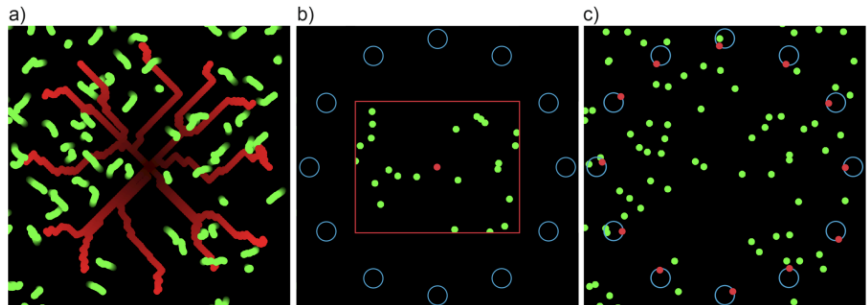

Fig. 6. The virtual gym environment showing trajectories to 12 target locations, subject to the same free microsphere motions. a) The trajectories chosen by the neural network (red) and the movements of the other microspheres (green). b) The start of all episodes, showing microsphere initial positions, target positions and the field of view permitted to the neural network. c) The entire environment at the end of the episodes.

## 6. Demonstration in the physical environment

Once the neural network was trained, it was applied to the physical environment where, importantly, no further training took place. An example trajectory is presented in Fig. 7, which shows the field of view (red rectangle) at time steps of a) 0, b) 74, and c) 146, corresponding to the initial, middle and end states. For visual clarity, the figures also show the environment state outside the field of view for positions already explored by the neural network (although the network has no mechanism to retain this information). As before, for each time step, the neural network was provided only with a binary image corresponding to the field of view, along with a vector denoting the direction towards the target position. The trajectory in the physical environment resulted in the transportation of the microsphere to the target position with only slight deviations from the ideal path, as shown by the small degree of movement in the left-direction in the time steps before reaching the target. The movement direction selected by the neural network (the velocity action) at each of the featured timesteps is shown by the red arrow.

As the X and Y movement stages were independently controlled, the fastest microsphere movement corresponded to the hypotenuse where both X and Y stages were at their maximum speeds. As such, for a vertical movement (i.e. Y axis only), the X movement velocity value had no effect on the vertical component of the speed of travel. Therefore, for a desired vertical movement, the neural network was able to move continuously in the Y axis at the maximum speed, whilst alternating between maximum positive and negative movement directions in the X axis (a technique familiar to many players of older computer games). This effect is observed in Fig. 7, where diagonal movements (i.e. a combination of X and Y stage movements) are shown by the red arrows. Although the neural network was found to use such diagonal movements extremely frequently, it is important to realize that the network could choose any possible value for each of the two stages (i.e. the action space is continuous) and a distribution of actions was observed, as opposed to discrete diagonal actions. Whilst the cause of this movement strategy is challenging to identify due to the nature of obfuscation in a neural network, we attribute this effect to the fact that this movement technique can increase the field of view, whilst having no negative effect on the movement speed.

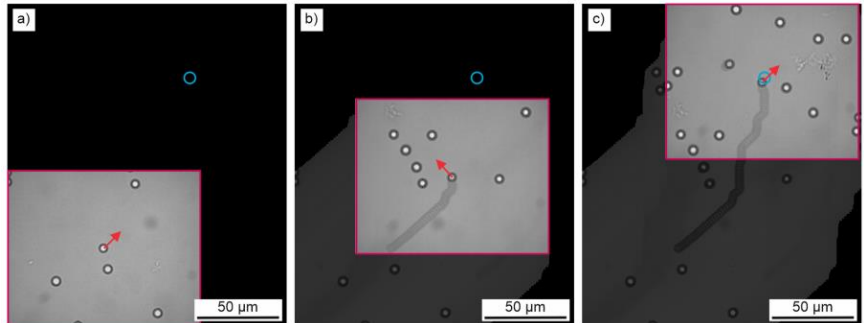

Fig. 7. Demonstration of the neural network in the physical environment, showing the field of view and trajectory path up to that point, at time steps of a) 0, b) 41, and c) 81.

## 7. Augmented physical environment

Training a neural network in a virtual environment for application in a physical environment is only possible if the neural network treats both environments equally. This process therefore has the interesting consequence that a combination of a virtual and a physical environment is also treated equally. This prospect is explored in this section where the capability to overlay a virtual environment onto the physical environment is demonstrated. Such a technique has the potential to unlock capabilities associated with mixed and augmented reality, such as adding safety limits for machine motion or using observation inputs from additional sensing techniques.

Fig. 8 shows a trajectory of a trapped microsphere controlled by the neural network in such an augmented physical environment, where there are both free-moving microspheres (that are physically present) and gate-shaped obstacles (that are virtually overlaid). (Note that stage velocity is higher than was used in Fig. 7 and so the distance moved per timestep is larger). The trajectory, which is displayed as a series of overlapping experimentally measured camera images, shows the successful movement of the laser-trapped microsphere to the opposite side of the environment. This is a clear demonstration that the neural network has learnt the general principles of obstacle avoidance and motion planning, as the neural network was trained on virtual free-moving microspheres but is shown here to be able to apply this learnt capability to a maze-like path formed of virtual microspheres. The virtual obstacles were rendered in the form of solid circles, equivalent to the appearance of the physical microspheres, to be treated equivalently to the free-moving physical microspheres. For visual clarity, the figure shows these virtual microspheres as white circles whilst the target position is show as a blue circle. The initial field of view of the microscope (the physical component of the augmented environment) is highlighted by the pink rectangle. In the augmented environment it is possible for the virtual components to be rendered over a wider field of view; this is the case here, with the augmented field of view illustrated by the green rectangle. The purple arrow shows the position where the environment has been split in two, for the purpose of displaying over two lines in the figure.

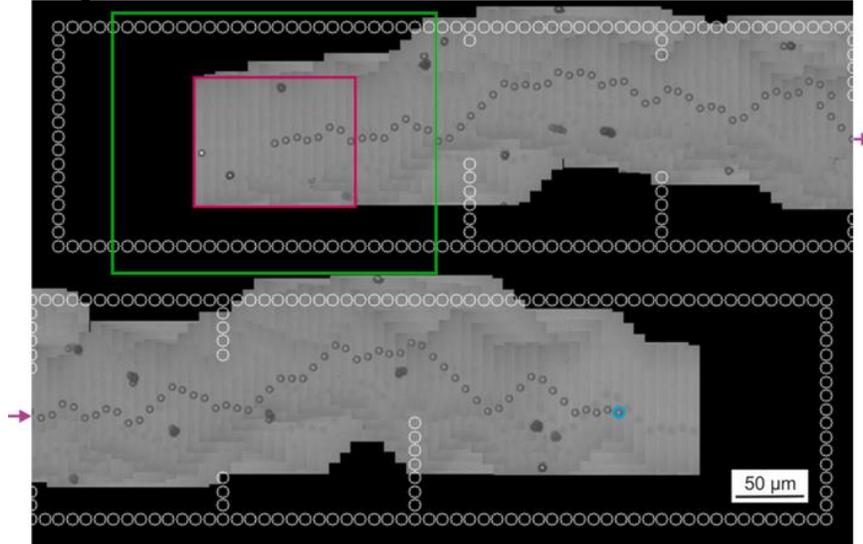

Fig. 8. Demonstration of neural network in the augmented physical environment, showing the virtual obstacles (white circles), the physical free-moving microspheres and the physical laser-trapped microsphere. The virtual obstacles are presented in the form of hollow circles in the figure for visual clarity. The solid rectangles represent the field of view of the physical microscope and the expanded field of view possible for the virtual environment.

## 8. Conclusions

Reinforcement learning offers the capability for a neural network to find a solution to a problem through self-exploration of the environment. However, in many cases, it is beneficial to train the neural network in a virtual environment, so that this exploration can occur faster and without the possibility of damage to the physical apparatus. However, this leads to the challenge, programmatically, of designing a sufficiently accurate simulation of the physical world. Here, an optical tweezers environment is designed, and a neural network is trained, with the objective of moving a microsphere to the desired position using the optical tweezers effect, whilst simultaneously avoiding collisions with other microspheres. The neural network, which was provided with limited field of view camera images and a vector indicating the direction towards the target, was shown to be able to achieve continuous control of the X,Y stage velocities and ensure that the laser-trapped microsphere reached the target destination without collision. The results presented demonstrate the effective transference of capability learned in the virtual environment to the physical environment and, furthermore, highlight the potential for applications in augmented environments, where a virtual environment was overlaid upon the physical one. The concept of training neural networks within virtual environments clearly has significant potential in the application of machine learning for experimental optimization and control. Fully unlocking this potential will require widespread application of these transformative techniques across a broad range of disciplines – a trend that is now entering the field of optics and photonics.

## Funding

Engineering and Physical Sciences Research Council (EP/N03368X/1).

## Acknowledgement

We gratefully acknowledge the support of NVIDIA Corporation with the donation of the Titan X GPU used for this research. The authors acknowledge the use of the IRIDIS High Performance Computing Facility, and associated support services at the University of Southampton, in the completion of this work.

## Disclosures

The authors declare no conflicts of interest.